\documentclass[10pt,conference,letterpaper]{ieeeconf}
\usepackage[utf8]{inputenc}
\usepackage{amsmath}
\usepackage{amssymb}
\usepackage{cite}
\usepackage{graphicx}
\usepackage{mathnotation}

\DeclareRobustCommand{\groupderiv}[1]{\accentset{\scriptstyle\circ}{#1}}
\DeclareRobustCommand*{\IEEEauthorrefmark}[1]{\raisebox{0pt}[0pt][0pt]{\textsuperscript{\footnotesize\ensuremath{\ifcase#1\or *\or \dagger\or \ddagger\or%
    \mathsection\or \mathparagraph\or \|\or **\or \dagger\dagger%
    \or \ddagger\ddagger \else\textsuperscript{\expandafter\romannumeral#1}\fi}}}}

\title{Characterizing Error in Noncommutative Geometric Gait Analysis}
\author{
    \authorblockN{
        Capprin Bass\IEEEauthorrefmark{1},
        Suresh Ramasamy\IEEEauthorrefmark{2}, and
        Ross L. Hatton\IEEEauthorrefmark{3}
    }
    \authorblockA{
        Collaborative Institute for Robotics and Intelligent Systems (CoRIS), Oregon State University\\
        Corvallis, Oregon \\
        Email: \IEEEauthorrefmark{1}basscap@oregonstate.edu, \IEEEauthorrefmark{2}sureshr14@gmail.com, \IEEEauthorrefmark{3}ross.hatton@oregonstate.edu
    }
}

\begin{document}

\maketitle

\begin{abstract}
    A key problem in robotic locomotion is in finding optimal shape changes to effectively displace systems through the world.
    Variational techniques for gait optimization require estimates of body displacement per gait cycle; however, these estimates introduce error due to unincluded high order terms.
    In this paper, we formulate existing estimates for displacement, and describe the contribution of low order terms to these estimates.
    We additionally describe the magnitude of higher (third) order effects, and identify that choice of body coordinate, gait diameter, and starting phase influence these effects.
    We demonstrate that variation of such parameters on two example systems (the differential drive car and Purcell swimmer) effectively manages third order contributions.
\end{abstract}

\section{Introduction}\label{sec:introduction}
In nature, creatures of all kinds move with gaits.
Bodies interact with their environment through changes in shape, which displace the body through the world.
These gaits are, by nature, cyclic: running, swimming, and flying all involve repeated action, and result in body displacement.
It is useful to describe the locomotion of robots in the same way.

When controlling locomoting robots, it is useful to understand which gait cycles result in ``good" displacements, based on desired gait properties such as displacement per unit time or unit energy \cite{ramasamy2016soap, hatton2021inertia}.
As one approach to this problem, the geometric mechanics community has described a framework for relating system dynamics, configuration, and gait geometry that provides insight into the displacements resulting from particular gaits \cite{Murray:1993,Morgansen:2007,walsh95,Mukherjee:1993a,Kelly:1995,Radford:1998,Melli:2006,Shammas:2007,Avron:2008}.

\begin{figure*}[t]
    \centering
    \includegraphics[width=0.85\linewidth]{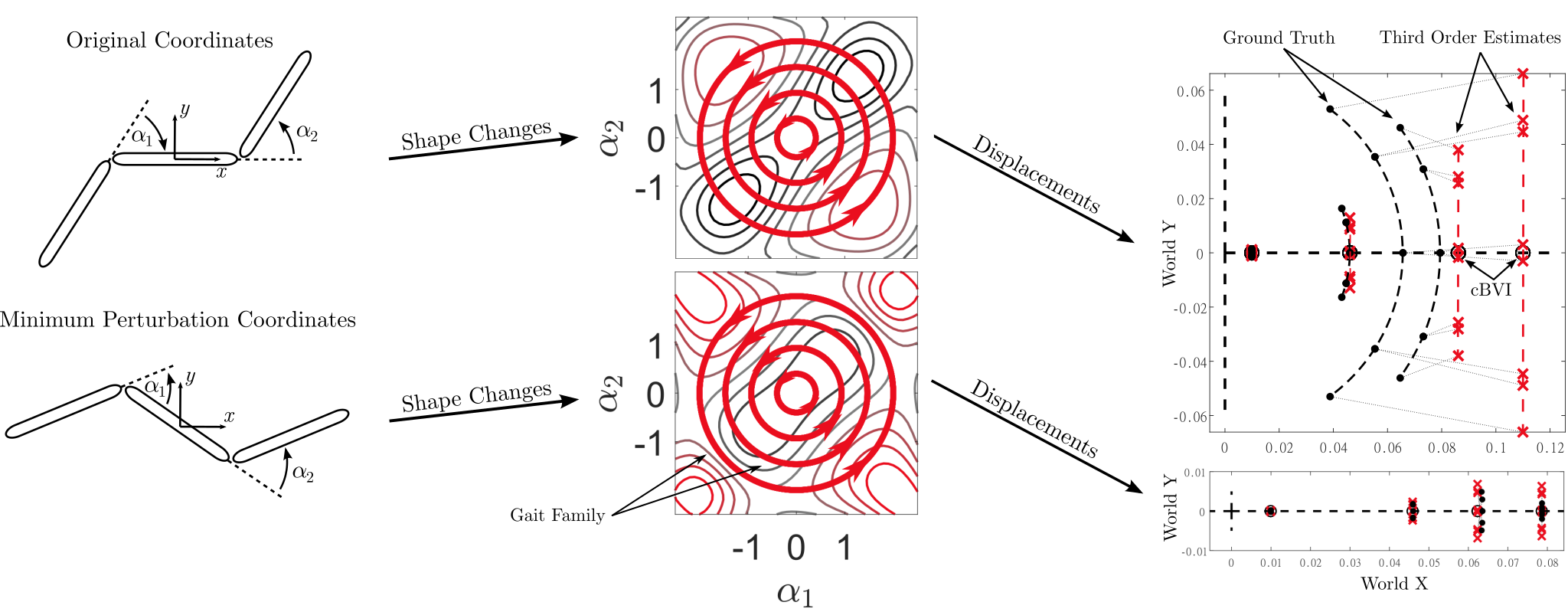}
    \caption{Relationship between systems and displacements. The configuration of a system is represented by shape variables; in this case, the Purcell swimmer is represented by two relative orientations $\alpha_1$ and $\alpha_2$. A \textit{gait family} captures shape changes of a certain kind. Integration of the BCH series approximation inside the gait provides estimates for displacement. As gait diameter increases, so does displacement per cycle; starting phase impacts gait error angle.}
    \label{fig:gm_span}
\end{figure*}

Because gaits are cycles in system shape, they form closed loops in the shape space of the system.
The \textit{motility map} $\mathbf{A}$, defined over the shape space of a system, can be used to map shape velocity to body velocity \cite{Radford:1998}.
Using the corrected body velocity integral (cBVI), a surface integral of the \textit{total Lie bracket} over the region enclosed by a gait, we construct an estimate for displacement, $g_\phi $ \cite{ramasamy2016soap, Hatton:2015EPJ},
\begin{equation}\label{eqn:tlb}
    g_\phi \approx \exp\Bigg{(}\underbrace{\iint_\phi \overbrace{d\mathbf{A} + [\mathbf{A}_1, \mathbf{A}_2]}^\text{total Lie bracket}}_\text{cBVI}\Bigg{)},
\end{equation}

\noindent in which the first term ($d\mathbf{A}$) captures the nonconservativity of locomotion; this is the ``forwards minus backwards" displacement due to the gait.
The second term ($[\mathbf{A}_1, \mathbf{A}_2]$) is the local Lie bracket of the matrix columns of the motility map,\footnote{
    The local Lie bracket may be extended to greater than two dimensions by taking Lie brackets of each matrix column: $\sum_{j>i} [\mathbf{A}_i,\mathbf{A}_j]$ \cite{ramasamy2016soap}.
} and captures the effects of noncommutativity of the position space; this is the sideways ``parallel parking" effect from ``move forward and turn" actions.
Fig. \ref{fig:gm_span} captures this relationship between system, shape changes, and estimated displacements for the Purcell swimmer \cite{Purcell:1977}.

Our previous work has shown that the cBVI has associated error; that is, the displacement predicted by the cBVI is not exactly ground truth.
This error comes from unaccounted-for higher order displacement effects present in many systems and gaits.
We have argued that particular choices of body coordinates (in particular, the use of minimum perturbation coordinates \cite{Hatton:2011IJRR}) reduce the contribution of higher order terms, instead capturing their effect with the total Lie bracket \cite{Hatton:2015EPJ}.
However, we have not previously quantified the error introduced by these higher order terms.

In this paper, we address the specific gap of understanding in the magnitude and direction of higher order terms of the total Lie bracket.
We do so by using the Baker-Campbell-Hausdorff series to construct an expression for the cBVI that includes higher order terms:
\begin{equation}\label{eqn:tlb_ext}
    \begin{split}
        g_\phi = \exp\Bigg{(}\iint_\phi&\overbrace{\bigl(d\mathbf{A} + [\mathbf{A}_1, \mathbf{A}_2]\bigr)}^\text{total Lie bracket} +\\ &\underbrace{\frac{\pi \ell}{8}\left[\bar{\mathbf{A}}, \iint_{\phi} \bigl(d\mathbf{A}+[\mathbf{A}_1,\mathbf{A}_2]\bigr)\right]}_\text{third order effects} + \cdots\Bigg{)},
    \end{split}
\end{equation}
in which $\bar{\mathbf{A}}$ is an estimate for the average of the motility map in the region of the gait and $\ell$ is the characteristic diameter of the gait in the shape space.

We comment further on the factors contributing to leading order error.
Third order effects (in the plane) are bounded by
\begin{equation}
    ||\mathbf{A}||\cdot||D\mathbf{A}|| \ell^3,
\end{equation}

\noindent where $D\mathbf{A}=d\mathbf{A} + [\mathbf{A}_1, \mathbf{A}_2]$, referring to to the total Lie bracket.
Because the cBVI is an area integral of $D\mathbf{A}$, third order effects may be expressed \textit{relatively} to the cBVI as being proportional to
\begin{equation}\label{eqn:rel}
    ||\mathbf{A}||\ell,
\end{equation}

\noindent the magnitude of the motility map and the size of the gait.
Given $||\mathbf{A}||$ for a system at a given point, (\ref{eqn:rel}) communicates the maximum size gait possible before error becomes too large.
Coordinate choice also affects $||\mathbf{A}||$; in fact, our own minimum perturbation coordinates maximize $\ell$ for a given level of acceptable error.
Fig. \ref{fig:gm_span} demonstrates this effect for two choices of coordinate and several gait amplitudes.

In addition, for gaits without net body rotation, third order effects are directed orthogonally to the displacement predicted by the cBVI.
As a result, their respective contributions may either be compared in terms of absolute magnitude or in terms of the ``error angle" resulting from third order effects.

The rest of the paper is organized as follows:
In \S\ref{sec:background}, we describe the model, as well as the supporting mathematics leading up to the total Lie bracket.
In \S\ref{sec:tlb}, we construct the total Lie bracket.
In \S\ref{sec:method}, we approximate third order effects, and derive expressions for third order contributions and a heuristic on characteristic gait diameter.
In \S\ref{sec:application}, we apply these methods to two locomoting systems.
In \S\ref{sec:conclusion}, we make concluding remarks and comment on future work.

\section{Model Background}\label{sec:background}

\subsection{Model}
We model our systems as having a configuration space $Q$, partitioned into a position space $G$ and a shape space $R$, as in \cite{hatton2021inertia}.
Elements $g \in G$ describe positions of the system in space, and $r \in R$ describe the shape of the system itself.
Fig. \ref{fig:gm_span} illustrates the difference between position and shape.
As in \cite{hatton2021inertia, Radford:1998, Hatton:2015EPJ, Hatton:2011IJRR}, the \textit{local connection} (or motility map) may be used to map infinitesimal shape changes to infinitesimal position changes,\footnote{
    This expression makes the assumption that systems behave \textit{kinematically}. Previous work \cite{revzen2019stokes} extends this domain to apply to many systems.
}
\begin{equation}\label{eqn:lc}
    \groupderiv{g} = \mathbf{A}(r) \dot{r},
\end{equation}
\noindent in which $\mathbf{A}$ refers to the local connection,\footnote{In previous work, the local connection, by convention, encodes negative body motion; we have dropped this convention for this paper.} and $\groupderiv{g}$ is a body velocity.
Body velocities are elements of the Lie algebra of the position space; they represent velocity in the local frame.
As such, Lie algebra elements may be represented by either a column vector (with the body frame acting as bases) or in a corresponding matrix form.
For the remainder of this paper, the position space is the special Euclidean group ($g \in SE(2)$); we notate body velocities with $\groupderiv{g} \in \mathfrak{se}(2)$.

\subsection{Gaits}
Certain changes in system shape result in a displacement through the position space.
In the context of locomotion, it is useful to describe shape changes in terms of  cyclic \textit{gaits}, where a mapping $\phi: [0,T) \to R$ describes the shape $r$ at time $t \in [0, T)$; $T$ is the period of the gait.
This structure allows us to express displacement from the identity induced by a gait:
\begin{equation}\label{eqn:disp}
    g_{\phi} = \int_0^T g(t) \mathbf{A}(r(t)) \dot{r}(t)dt = \ointctrclockwise_\phi g\mathbf{A}(r) dr,
\end{equation}
\noindent where the rightmost integral described in (\ref{eqn:disp}) is a path integral along a closed loop drawn in the shape space by the gait $\phi$.

This integral is invariant to time parameterizations, but does depend on the ordering of actions along the path.
Both versions of the integral contain system configuration, which a traditional Riemann integral does not adequately describe.
In contrast, the product-integral\footnote{
    The product-integral is a \textit{multiplicative} version of the additive Riemann integral.
    Product integration preserves the effect of the group operation, rather than integrating components independently.
    In effect, the product integral preserves the order that events occur, as in (\ref{eqn:disp}).
}
accounts for the ordering of actions along the path, respecting configuration:
\begin{equation}\label{eqn:prodi}
    g_\phi = \Prodi_0^T \Big{(}\exp\bigl(\mathbf{A}(t)\dot{r}(t) dt\bigr)\Big{)},
\end{equation}
in which exponentiating $\mathbf{A}(t)\dot{r}(t)$ over infinitesimal time produces the corresponding body frame transformation, and taking the product of all these infinitesimal transformations produces the total displacement over the gait.

Because multiplication of translation/rotation elements does not commute, we still cannot compute a closed form expression for this integral. By employing the Baker-Campbell-Hausdorff series described in the next section, however, we can construct an approximate closed-form solution that provides geometric insight into the system motion.

\subsection{The Baker-Campbell-Hausdorff Series}\label{sec:bch}
The Baker-Campbell-Hausdorff (BCH) series expresses the result of executing serial group actions as a single equivalent operation.
It is related to the exponential map, which implies a correspondence between a groupwise velocity and a group action.
For example, take a groupwise velocity $\groupderiv{g} \in \mathfrak{se}(2)$; it is mapped to a group element $g \in SE(2)$ with:
\begin{equation}
    g = \exp\left(\groupderiv{g}\right),
\end{equation}

\noindent where the exponential map is equivalent to \textit{integration} of the groupwise velocity over unit time.

Now, take two groupwise velocities $X, Y \in \mathfrak{se}(2)$.
Applying their corresponding group actions in series has the form
\begin{equation}\label{eqn:exps}
    g = \exp\left(X\right) \exp\left(Y\right).
\end{equation}

\noindent The BCH series can be used to replace the right hand side of (\ref{eqn:exps}) with the exponential of a single groupwise velocity $Z \in \mathfrak{se}(2)$ defined such that
\begin{equation}
    \exp(Z) = \exp(X)\exp(Y).
\end{equation}

\noindent The BCH series is infinite, and its lowest order terms are:
\begin{equation}
    Z = X + Y + \frac{1}{2}[X,Y] + \frac{1}{12}[X-Y,[X,Y]] + \cdots.
\end{equation}

\begin{figure}[t]
    \centering
    \includegraphics[width=\linewidth]{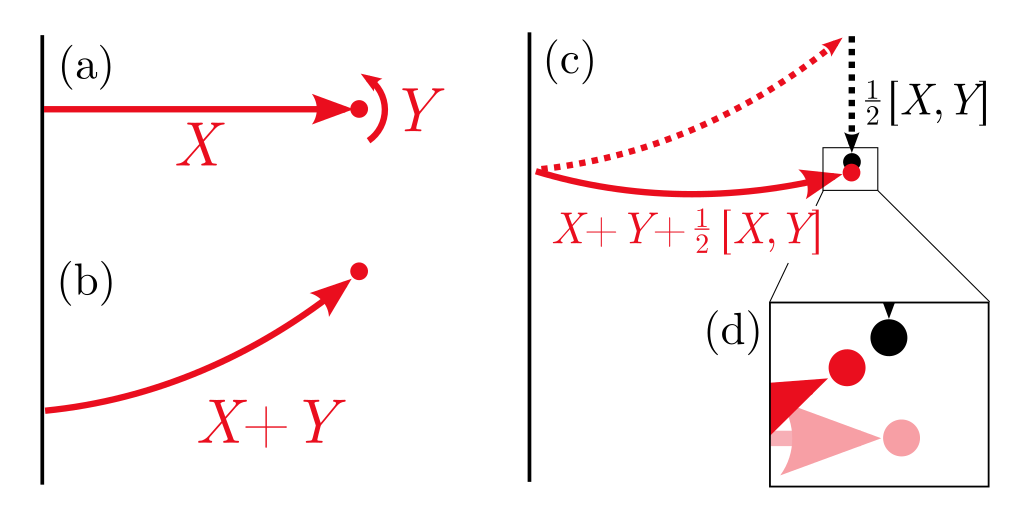}
    \caption{Action of the BCH series. (a) For the serial actions $X$, $Y$, the ground truth is moving forward, followed by a turn. (b) The action $X+Y$ results in movement along an arc. (c) The inclusion of a local Lie bracket term, $\frac{1}{2}[X,Y]$, corrects some error. (d) The endpoints for ground truth (pink), Lie bracket correction (black), and truncated BCH series (red) are distinct.}
    \label{fig:bch}
\end{figure}

\noindent Note that the BCH series contains the nominal $X + Y$ as expected from commutative algebra; however, it also contains additional, corrective terms.
The following example builds intuition for these terms; refer to Fig. \ref{fig:bch} for its visualization.

Consider the example of a diffdrive car, which can \textit{drive forward} and \textit{turn}.
We assign $X$ as driving forward, and $Y$ as turning; because $X,Y \in \mathfrak{se}(2)$, we write each action as
\begin{equation}
    X = [\groupderiv{x}\quad 0 \quad 0]^T,
    \quad
    Y = [0 \quad 0 \quad \groupderiv{\theta}]^T,
\end{equation}

\noindent for some $\groupderiv{x}, \groupderiv{\theta}$.
The composite motion $\exp(X)\exp(Y)$ encodes displacement after driving forward for some time, and then turning.
This results in a $(x, 0, \theta)$ position.
In contrast, $\exp(X+Y)$ encodes displacement after simultaneously driving forward \textit{and} turning, resulting in a $(x,y,\theta)$ position.

The two operations result in different predicted displacements of the car in space; $\exp(X)\exp(Y)$ is the ground truth, and $\exp(X+Y)$ is an approximation of ground truth, discarding information about the order in which events occur.
To improve the approximation, we can introduce additional terms from the BCH series.
The second term expands the approximation to $\exp(X+Y+\frac{1}{2}[X,Y])$.
The local Lie bracket captures the fact that $X$ occurred \textit{before} $Y$, and  introduces a lateral velocity to the car that corrects most of the $y$ error in the approximation.

\section{Baker-Campbell-Hausdorff for Gaits}\label{sec:tlb}

Rather than integrating (\ref{eqn:disp}) or (\ref{eqn:prodi}) directly, we construct an integral estimate that captures the relationship between system properties and displacement.
This estimate leverages the BCH series to describe leading-order displacement effects from gaits, while simplifying the integral expression such that it may be solved numerically.
The total Lie bracket is a truncation of this BCH series expression, as we will show.

\begin{figure}[tb]
    \centering
    \includegraphics[width=0.55\linewidth]{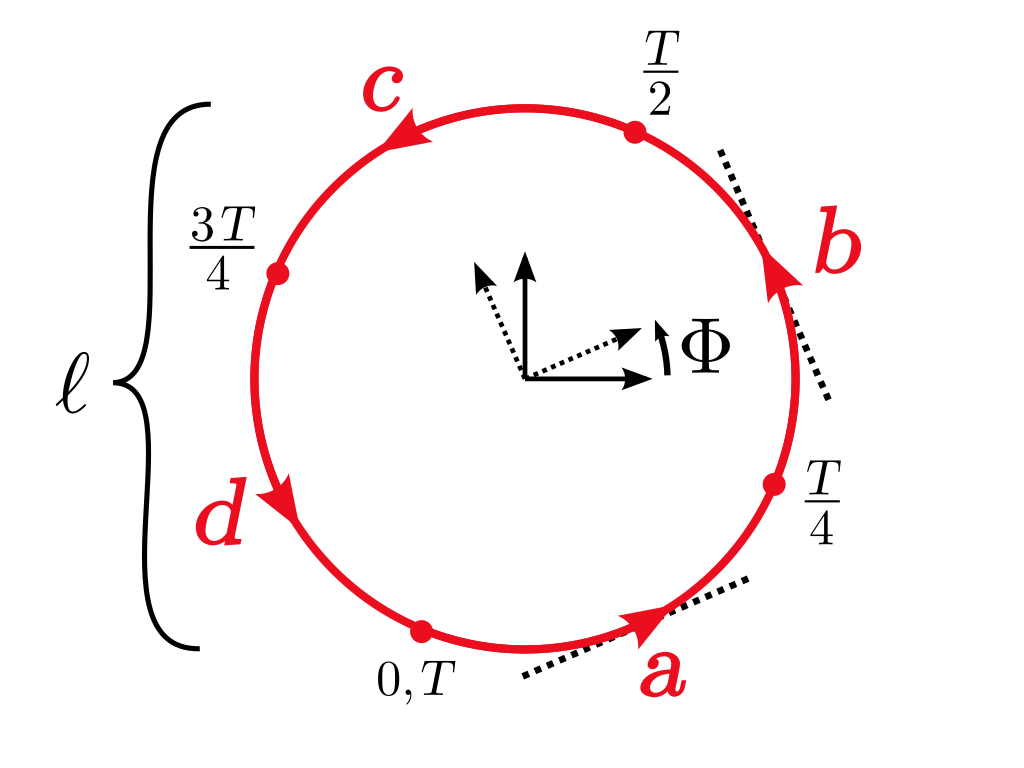}
    \caption{To construct a displacement integral, a circular gait of diameter $\ell$ and starting phase $\Phi$ is discretized into four segments $a$, $b$, $c$, $d$.}
    \label{fig:gait_split}
\end{figure}

We first split the gait into four sections $a$--$d$, distributed evenly around the gait such that the mean tangent vectors in $a$ and $c$ are antiparallel, as are the mean tangent vectors in $b$ and $d$, as illustrated in Fig.~\ref{fig:gait_split}. This split discretizes the product-integral from~\eqref{eqn:prodi} into the product of four product-integrals over smaller intervals,

\begin{equation}\label{eqn:disp4}
    g_\phi \approx \prod_{i=1}^{4} \left(\Prodi_{(i-1)T/4}^{(i)T/4} \Big{(}\exp\bigl(\mathbf{A}(t)\dot{r}(t) dt\bigr)\Big{)}\right) 
\end{equation}

If we apply the BCH series recursively to the (infinite number of infinitesimal) elements in each of the four product integrals and assume commutativity within each gait segment such that all the Lie bracket terms go to zero, the segment integrals may be written as
\begin{equation}
    \Prodi_\tau\Big{(}\exp\bigl(\mathbf{A}(t)\dot{r}(t) dt\bigr)\Big{)} = \exp\left(\int_\tau \mathbf{A} d\tau\right),
\end{equation}
where $\tau$ is an arbitrary gait segment.\footnote{As per~\cite{ramasamy:thesis}, the assumption of local commutativity introduces fourth-order errors; we constrain our focus in this paper to third-order errors.} 

Assigning each of the $\int \mathbf{A}$ integrals to their corresponding segment names, we can rewrite (\ref{eqn:disp4}) as the product of four exponential terms,
\begin{equation}
    g_\phi \approx e^a e^b e^c e^d.
\end{equation}
Applying the BCH formula to each term in this product produces a series expression for $g_{\phi}$ in terms of the $\int \mathbf{A}$ integrals,
\begin{equation}\label{eqn:disp1}
    \begin{split}
        \approx \exp(a + b + c + d + \frac{1}{2}([a,b] + [b,c] 
         + \cdots,
    \end{split}
\end{equation}
in which the first terms are the ``simple integral" of the body frame motions the system makes, and the Lie bracket terms are corrections to the global motion based on the order in which the segments appear in the gait.

With the series expression from \eqref{eqn:disp1} in hand, we can now use the geometric arrangement of the segments to gain further insight about the gait displacement integral: Because the elements of the gait pairs $\{a, c\}$ and $\{b, d\}$ are antiparallel, we can approximate them in terms of the mean value of the local connection in the region of the gait, its first derivative over the shape space, and the characteristic diameter $\ell$ of the gait in the shape space. This approximation takes the form
\begin{equation}\label{eqn:params}
    \begin{aligned}
        a &\approx \alpha - \delta/2, &
        b &\approx \beta + \Delta/2,\\
        c &\approx -(\alpha + \delta/2), &
        d &\approx -(\beta - \Delta/2),
    \end{aligned}
\end{equation}
in which $\alpha$ and $\beta$ are the mean values of the columns of the local connection in directions aligned with the $a$ and $b$ sections of the gait, scaled by $\ell\pi /4$, the length of a quarter-circle for diameter $\ell$,
\begin{equation}\label{eqn:ab}
    \alpha = \frac{\pi}{4} \ell \bar{\mathbf{A}}R(\Phi)
    \begin{bmatrix}
        1\\0
    \end{bmatrix},
    \quad
    \beta = \frac{\pi}{4} \ell \bar{\mathbf{A}}R\left(\Phi + \frac{\pi}{2}\right)
    \begin{bmatrix}
        1\\0
    \end{bmatrix},
\end{equation}
and $\delta$ and $\Delta$ are the rates at which the local connection changes across the shape space, multiplied by the diameter of the gait, 
\begin{equation}
    \delta = \ell \frac{\partial \mathbf{A}_\alpha}{\partial r_\beta},
    \quad
    \Delta = \ell \frac{\partial \mathbf{A}_\beta}{\partial r_\alpha}.
\end{equation}

Inserting these approximations into the BCH series for the gait gives
\begin{equation}\label{eqn:disp_to}
    \begin{split}
        g_\phi \approx \exp(&-\delta + \Delta + [\alpha, \beta]+\\
        &\frac{1}{2}[(\alpha+\beta), (-\delta + \Delta + [\alpha, \beta])] +\cdots,
    \end{split}
\end{equation}
which we can then expand in terms of the local connection as
\begin{equation}\label{eqn:disp_full}
    \begin{split}
        g_\phi &\approx
        \exp \left( \underbrace{
            \overbrace{\iint_\phi \bigl(d\mathbf{A}}^{-\delta + \Delta} +
            \overbrace{\rule{0pt}{14pt}[\mathbf{A}_1, \mathbf{A}_2]\bigr)}^{[\alpha, \beta]}
        }_{\text{cBVI}} \right. +\\
        &\phantom{\exp()()}
        \left.\overbrace{\frac{\pi \ell}{8}\left[(\bar{\mathbf{A}}_{\alpha}+\bar{\mathbf{A}}_{\beta}), \iint_{\phi} \bigl( d\mathbf{A}+[\mathbf{A}_1,\mathbf{A}_2]\bigr)\right]}^\text{third order effects} \right).
    \end{split}
\end{equation}

This surface integral formulation opens the possibility of gait optimization via variational techniques.
These specific optimization techniques are outside the scope of this paper; however, they assume that the cBVI is an accurate estimate of displacement (as in the optimized coordinates shown in the last row of Fig. \ref{fig:gm_span}).
Our focus in this paper is the validity of this assumption, and quantifying residual errors due to the truncation of the BCH series.
In particular, we express and bound the contribution of third order effects to displacement.

\section{Third Order Bound}\label{sec:method}
The third order effects in (\ref{eqn:disp_full}) depend on $(\alpha + \beta)$ and the cBVI.
As in (\ref{eqn:ab}), the gait diameter $\ell$ and starting phase $\Phi$ together encode the initial configuration of the system.
Within a fixed diameter gait, $\Phi$ is solely responsible for initial system configuration.
In general, third order effects depend on the size and orientation of the local connection $\mathbf{A}$, the gait diameter $\ell$, and starting phase $\Phi$, as shown in Fig. \ref{fig:gm_span}.

\subsection{Approximation}
Third order effects (\ref{eqn:disp_full}) are a third-degree polynomial in $\ell$; to produce a third order bound, we construct a similar polynomial approximation for the cBVI.
This is done by computing a two-dimensional, second order Taylor series approximation for the total Lie bracket, $D\mathbf{A}$, at the center of the gait:
\begin{equation}\label{eqn:tlb_est}
    D\mathbf{A} \approx \overline{D\mathbf{A}}(\delta\alpha_1, \delta\alpha_2) = T_{D\mathbf{A}}^2(\delta\alpha_1, \delta\alpha_2).
\end{equation}
We then reparameterize (\ref{eqn:tlb_est}) into polar coordinates, and integrate over the circular approximation for a gait,
\begin{equation}
    \overline{\text{cBVI}}(\ell) = \int_0^{\ell/2} \int_0^{2\pi} \overline{D\mathbf{A}}(\rho, \theta) \rho\: d\rho\: d\theta,
\end{equation}
\noindent producing a diameter-dependent estimate for the cBVI.

As third order effects (and associated estimates) depend on phase $\Phi$ and characteristic gait diameter $\ell$, these parameters are used to compute a third order bound.

\begin{figure*}[t]
    \centering
    \includegraphics[width=0.85\linewidth]{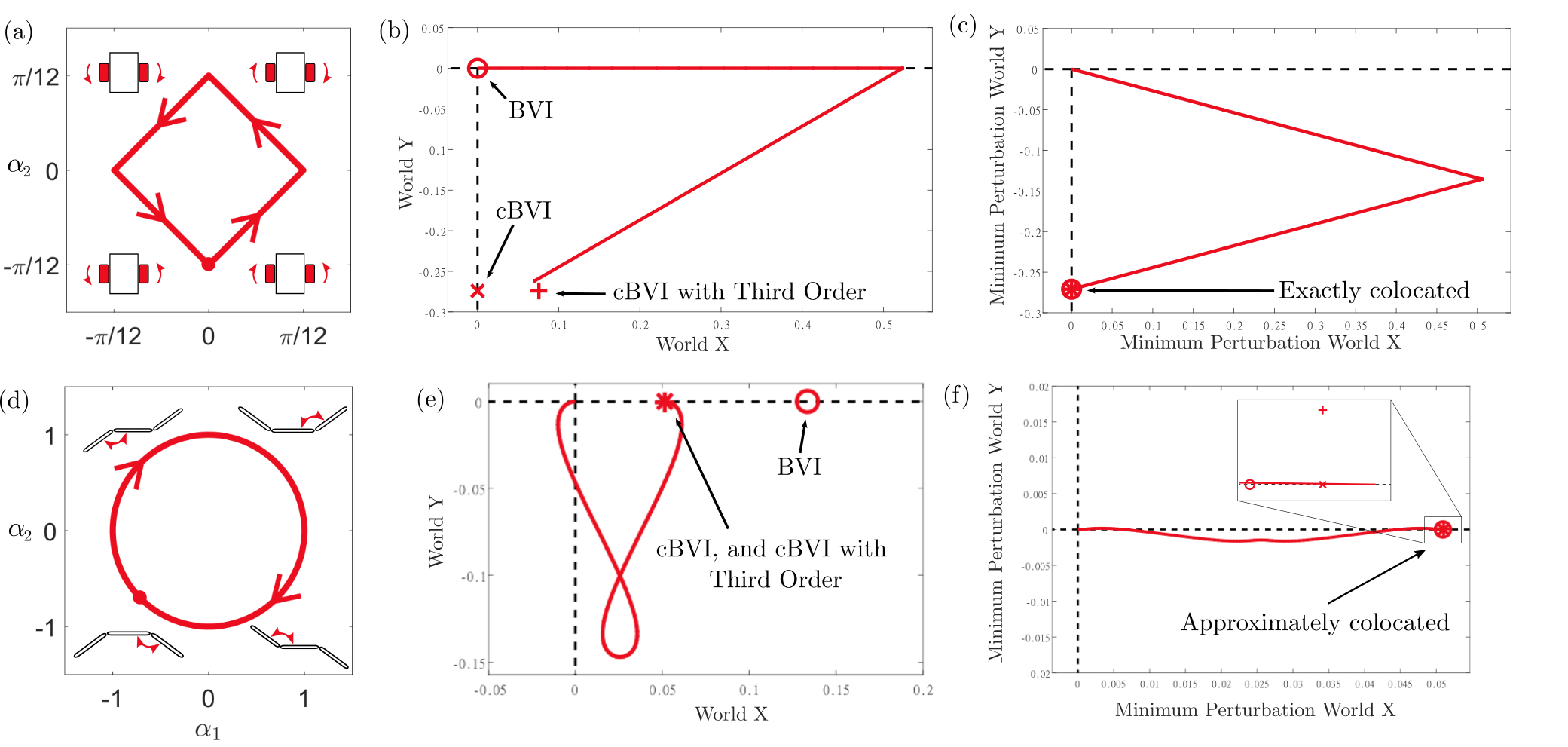}
    \caption{Characteristic gaits and resulting trajectories for each system. On the top row, (a-c), the diffdrive car moves in square gaits (a), with the orientation of each wheel as the shape. (b) Square diffdrive gaits result in a displacement trajectory. (c) In minimum perturbation coordinates, BCH estimates for displacement are exactly colocated. On the bottom row (d-f), the Purcell swimmer moves with circular gaits (d), with the relative orientation of each link as the shape. (e) Circular Purcell gaits result in a displacement trajectory. (f) In minimum perturbation coordinates, BCH estimates are only approximately colocated.}
    \label{fig:gaits}
\end{figure*}

\subsection{Third Order Bound}
The preceding polynomial approximations assume a circular gait, and express both nominal and approximate displacements as a result.
We use these approximations to construct a heuristic on the size of third order contributions for generic systems, and determine third order effects in the worst case.

The magnitude of third order effects can be made large by maximizing the possible size of its constituent components.
This is first done with the triangle inequality on $\alpha$ and $\beta$, creating an upper bound on $(\alpha + \beta)$:
\begin{equation}
    (\alpha + \beta) \leq (\abs{\alpha} + \abs{\beta});\quad
    (\abs{\alpha} + \abs{\beta}) = \frac{\pi}{4} \ell \abs{\bar{\mathbf{A}}}\begin{bmatrix}1\\1\end{bmatrix}.
\end{equation}
Note that with the absolute values of $\alpha$ and $\beta$, phase $\Phi$ is no longer present.
This implies that the bound captures the largest possible third order effects across all phases.

We also maximize the local Lie bracket, using the triangle inequality. This has the form
\begin{equation}
    [X, Y] =
    \begin{bmatrix}
        X^y Y^\theta - Y^y X^\theta\\
        Y^x X^\theta - X^x Y^\theta\\
        0
    \end{bmatrix}
    \leq
    \begin{bmatrix}
        \abs{X^y Y^\theta} + \abs{Y^y X^\theta}\\
        \abs{Y^x X^\theta} + \abs{X^x Y^\theta}\\
        0
    \end{bmatrix}.
\end{equation}
Combining the upper bounds, we have
\begin{equation}\label{eqn:fullbnd}
    \text{third order bound} = \frac{1}{2}[(\abs{\alpha} + \abs{\beta}), \overline{\text{cBVI}}(\ell)]_\text{ub},
\end{equation}
where $[\cdot, \cdot]_\text{ub}$ refers to the bound on the local Lie bracket.

An important note is that the local Lie bracket in (\ref{eqn:fullbnd}) speaks to the direction of third order effects.
For gaits with no net rotation, third order effects are orthogonal to the cBVI; this is demonstrated in Fig. \ref{fig:gm_span} and Fig. \ref{fig:sweep_orig}.
As a result, we may speak of third order effects in terms of the ``error angle" they produce.
Ground truth displacements lie on an arc with an equivalent angle, and a radius equivalent to the cBVI.
Location on the arc is determined by the starting phase $\Phi$ of the particular gait.

\subsection{Characteristic Length Bound}\label{sec:bound}
The third order bound is an increasing function of the characteristic diameter $\ell$ and the local connection $\mathbf{A}$; third order contributions are small if $\ell$ is small.
The definition of ``small" is relative, and is determined by the size of the local connection, which depends on the choice of body frame.
For a given choice of coordinates, solving the inequality
\begin{equation}\label{eqn:bound}
    [(\abs{\alpha} + \abs{\beta}), \overline{\text{cBVI}}(\ell)]_\text{ub} \leq P \cdot \overline{\text{cBVI}}(\ell)
\end{equation}
\noindent for $\ell$ will constrain third order effects to a proportion $P$ of the cBVI.
Because all the quantities involved are polynomials in $\ell$, (\ref{eqn:bound}) can be solved numerically.

\section{Application of Bound}\label{sec:application}

We now apply the techniques introduced in \S\ref{sec:method} for two systems, investigating particular gait families.
We explore the direction and magnitude of third order effects in both original and minimum perturbation coordinates.

\subsection{Systems}
We investigate two example systems: the differential-drive car and the Purcell swimmer.
Illustrations of each are in Fig. \ref{fig:gaits}.
Both reside in the plane and have two shape variables.
For the car, the shape variables are the orientations of the wheels; for the swimmer, they are the relative orientations of every two links.
A ``shape" is a particular value for both shape variables; it defines the configuration of a system.

In general, the shape space of a system is all of the possible shapes it can make; gaits are closed loops within the shape space.
In the case of the above systems, we can represent the shape space as a subset of $\mathbb{R}^2$, and draw closed loops on the plane to construct gaits of interest.

Fig. \ref{fig:gaits}(a) and \ref{fig:gaits}(d) show characteristic gaits investigated for each system. The diffdrive car has a square gait, as it executes discrete ``move forward" and ``turn" actions.
The Purcell swimmer has a circular gait, where it continuously accelerates each joint.
Each gait results in a displacement through the world; this displacement (and how it varies for different gaits) is of principal interest.

\subsection{Displacement and Effect of Additional Terms}
Changes in system shape induce a displacement trajectory through position space, shown in Fig. \ref{fig:gaits}(b) and \ref{fig:gaits}(e).

As mentioned in \S\ref{sec:background}, the ground truth is calculated exactly with a path integral of the local connection along the gait.
Approximations of displacement are done with a surface integral of BCH terms inside the path; as more terms are included, the approximation becomes more accurate.

\subsection{Minimum Perturbation Coordinates}

Choice of body coordinate affects the trajectory that systems follow through position space.
Body displacements may be computed in any body frame, so long as the frame is rigidly attached to the system.
Minimum perturbation coordinates \cite{Hatton:2011IJRR} are a choice of frame with this property.

Fig. \ref{fig:gaits}(c) and \ref{fig:gaits}(f) show how the use of minimum perturbation coordinates affects each system's trajectory.
For the diffdrive car, displacement estimates are exact in minimum perturbation coordinates \cite{ Hatton:2015EPJ, Hatton:2011IJRR}: the BVI,\footnote{The Body Velocity Integral (BVI) \cite{Hatton:2011IJRR} is a first order estimate for displacement, making Fig. \ref{fig:gaits} capture first, second, and third order estimates.} cBVI, and third order estimates are perfectly colocated.
For the Purcell swimmer, the BVI, cBVI, and third order estimates are approximately colocated; in this case, they are $0.0031$, $0.0013$, and $0.0022$ from ground truth, respectively.

It is important to note that Fig. \ref{fig:gaits} ignores the effect of starting phase on displacement, which does appear in higher order terms.
This effect is addressed in both \S\ref{sec:thirdorder} and \S\ref{sec:bound_app}.

\subsection{Relative Third Order Contribution}\label{sec:thirdorder}
To demonstrate the third order bound, we sample the gait families for both the diffdrive car and Purcell swimmer over gait amplitude and starting phase.
In addition, we compute resulting ground truth, cBVI, and third order effects for each gait, in original and minimum perturbation coordinates.

\begin{figure}[tb]
    \centering
    \includegraphics[width=\linewidth]{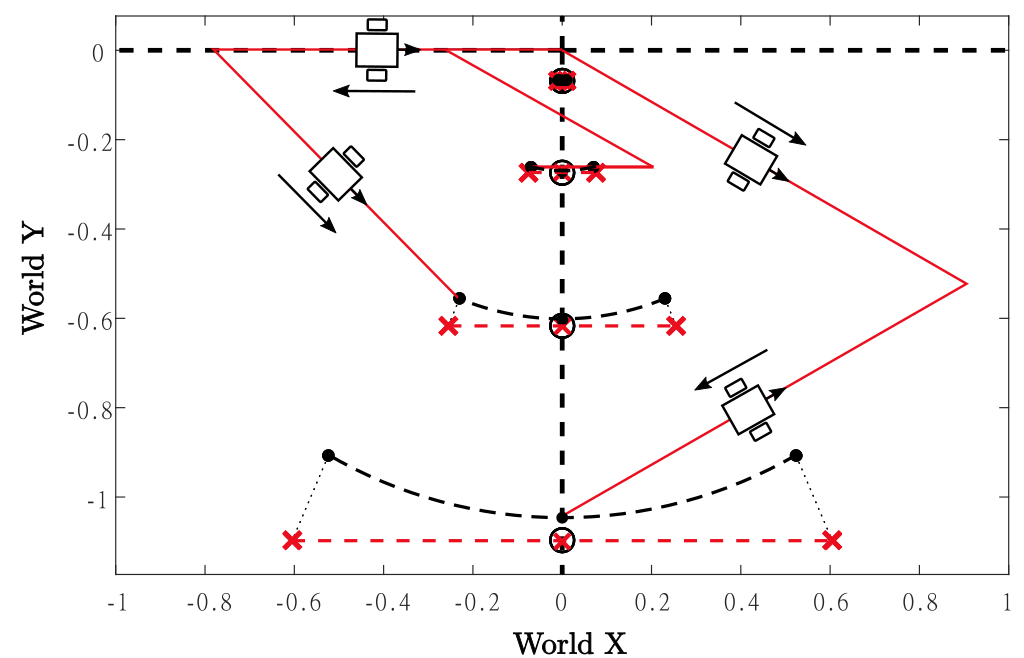}
    \caption{Amplitude and phase sampling for the diffdrive car, in original coordinates. Sample trajectories are included in red; resulting ground truths are in black. The cBVI captures changes in amplitude; third order effects (as red X's) capture changes in phase. Arc ``error" angle increases with amplitude.}
    \label{fig:sweep_orig}
\end{figure}

Fig. \ref{fig:sweep_orig} shows the results of amplitude and period sampling for the diffdrive car, in original coordinates.
The sampling is omitted in minimum perturbation coordinates, as the BVI exactly captures displacement for this system \cite{ Hatton:2015EPJ, Hatton:2011IJRR}.
Fig. \ref{fig:gm_span} shows the same sampling for the Purcell swimmer, in both original and minimum perturbation coordinates.

For both systems, it is clear that the magnitude of third order effects is much smaller than the cBVI, i.e., third order contributions to displacement are small (in optimal coordinates).
The relative \textit{size} of this contribution increases with gait amplitude.
In cases where the bound is not acceptably small, it can be made so with the constraints on amplitude.

\subsection{Guarantees using Third Order Bounds}\label{sec:bound_app}
\begin{figure}[tb]
    \centering
    \includegraphics[width=\linewidth]{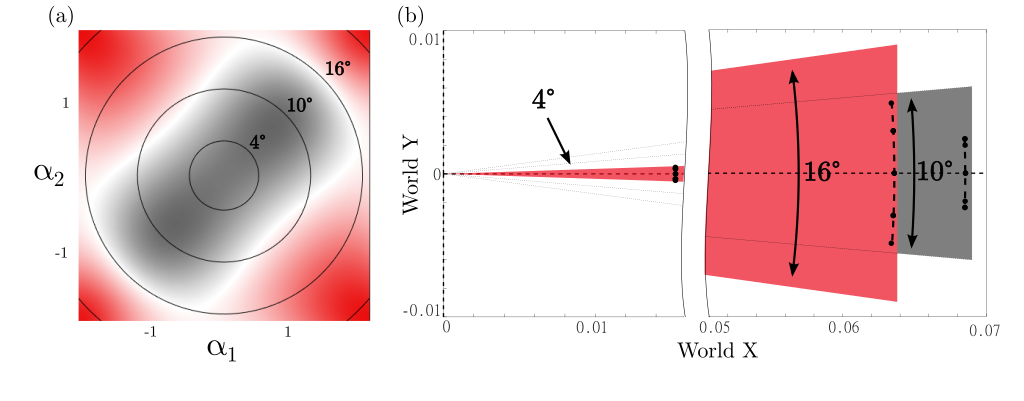}
    \caption{Error angle for the Purcell swimmer, in minimum perturbation coordinates. The X constraint curvature function is overlaid with gait contours; each gait diameter has an associated error angle (a). Error angle bounds the third order effects present in the ground truth (b). Ground truths (dashed lines) shown are of the same scale as in Fig. \ref{fig:gm_span}.}
    \label{fig:gait_bound}
\end{figure}

Using the length bound defined in \S\ref{sec:bound}, the magnitude of third order effects may be \textit{absolutely} constrained to an arbitrary proportion of the cBVI. The effect of this bounding technique is shown in Fig. \ref{fig:gait_bound}.
The maximum ``error angle" increases with characteristic gait diameter.

Third order effects may additionally be \textit{relatively} constrained (within a given amplitude) with an intelligent selection of starting phase, $\Phi$.
As shown in (\ref{eqn:disp_full}), the local Lie bracket increases with $(\alpha + \beta)$; $\alpha$ and $\beta$ are dependent on \textit{phase}, as well as characteristic diameter.
Certain choices of starting phase will result in relatively small third order effects.
Fig. \ref{fig:gm_span} and Fig. \ref{fig:sweep_orig} corroborate this claim.
Within a given amplitude, the sampled phases all have different third order contributions.

\section{Conclusion}\label{sec:conclusion}
In this paper, we extend existing displacement approximations, and characterize third order effects of the BCH series in the context of locomoting systems.
We identify that gait diameter, starting phase, and coordinate choice influence third order contributions, and demonstrate the use of these quantities as tools to manage errors introduced by the cBVI.

Future work will explore third order effects in the context of gaits with net rotation; these will act non-orthogonally to the cBVI, and require further analysis.
In addition, we will expand scope to include fourth order terms, which capture additional, previously ignored displacement effects.

\section*{Acknowledgment}
This work was supported in part by the National Science Foundation under grants 1653220 and 1826446.

\bibliographystyle{./IEEEtran}
\bibliography{./IEEEabrv,./rossbib}

\begin{thebibliography}{10}
\providecommand{\url}[1]{#1}
\csname url@samestyle\endcsname
\providecommand{\newblock}{\relax}
\providecommand{\bibinfo}[2]{#2}
\providecommand{\BIBentrySTDinterwordspacing}{\spaceskip=0pt\relax}
\providecommand{\BIBentryALTinterwordstretchfactor}{4}
\providecommand{\BIBentryALTinterwordspacing}{\spaceskip=\fontdimen2\font plus
\BIBentryALTinterwordstretchfactor\fontdimen3\font minus
  \fontdimen4\font\relax}
\providecommand{\BIBforeignlanguage}[2]{{%
\expandafter\ifx\csname l@#1\endcsname\relax
\typeout{** WARNING: IEEEtran.bst: No hyphenation pattern has been}%
\typeout{** loaded for the language `#1'. Using the pattern for}%
\typeout{** the default language instead.}%
\else
\language=\csname l@#1\endcsname
\fi
#2}}
\providecommand{\BIBdecl}{\relax}
\BIBdecl

\bibitem{ramasamy2016soap}
S.~Ramasamy and R.~L. Hatton, ``Soap-bubble optimization of gaits,'' in
  \emph{Decision and Control (CDC), 2016 IEEE 55th Conference on}.\hskip 1em
  plus 0.5em minus 0.4em\relax IEEE, 2016, pp. 1056--1062.

\bibitem{hatton2021inertia}
R.~L. Hatton, Z.~Brock, S.~Chen, H.~Choset, H.~Faraji, R.~Fu, N.~Justus, and
  S.~Ramasamy, ``The geometry of optimal gaits for inertia-dominated kinematic
  systems,'' 2021.

\bibitem{Murray:1993}
R.~M. Murray and S.~S. Sastry, ``Nonholonomic motion planning: Steering using
  sinusoids,'' \emph{IEEE Transactions on Automatic Control}, vol.~38, no.~5,
  pp. 700--716, Jan 1993.

\bibitem{Morgansen:2007}
\BIBentryALTinterwordspacing
K.~A. Morgansen, B.~I. Triplett, and D.~J. Klein, ``Geometric methods for
  modeling and control of free-swimming fin-actuated underwater vehicles,''
  \emph{IEEE Transactions on Robotics}, vol.~23, no.~6, pp. 1184--1199, Jan
  2007. [Online]. Available:
  \url{http://ieeexplore.ieee.org/xpls/abs{\_}all.jsp?arnumber=4399955}
\BIBentrySTDinterwordspacing

\bibitem{walsh95}
G.~C. Walsh and S.~Sastry, ``On reorienting linked rigid bodies using internal
  motions,'' \emph{Robotics and Automation, IEEE Transactions on}, vol.~11,
  no.~1, pp. 139--146, January 1995.

\bibitem{Mukherjee:1993a}
R.~Mukherjee and D.~P. Anderson, ``A surface integral approach to the motion
  planning of nonholonomic systems,'' in \emph{American Control Conference,
  1993}, 1993, pp. 1816 --1823.

\bibitem{Kelly:1995}
S.~D.~K. Kelly and R.~M. Murray, ``Geometric phases and robotic locomotion,''
  \emph{J. Robotic Systems}, vol.~12, no.~6, pp. 417--431, Jan 1995.

\bibitem{Radford:1998}
J.~E. Radford and J.~W. Burdick, ``Local motion planning for nonholonomic
  control systems evolving on principal bundles,'' in \emph{{Proceedings of the
  International Symposium on Mathematical Theory of Networks and Systems}},
  Padova, Italy, 1998.

\bibitem{Melli:2006}
J.~B. Melli, C.~W. Rowley, and D.~S. Rufat, ``Motion planning for an
  articulated body in a perfect planar fluid,'' \emph{SIAM Journal of Applied
  Dynamical Systems}, vol.~5, no.~4, pp. 650--669, November 2006.

\bibitem{Shammas:2007}
E.~A. Shammas, H.~Choset, and A.~A. Rizzi, ``Geometric motion planning analysis
  for two classes of underactuated mechanical systems,'' \emph{Int. J. of
  Robotics Research}, vol.~26, no.~10, pp. 1043--1073, 2007.

\bibitem{Avron:2008}
J.~E. Avron and O.~Raz, ``A geometric theory of swimming: {P}urcell's swimmer
  and its symmetrized cousin,'' \emph{New Journal of Physics}, vol.~9, no. 437,
  2008.

\bibitem{Hatton:2015EPJ}
R.~L. Hatton and H.~Choset, ``Nonconservativity and noncommutativity in
  locomotion,'' \emph{European Physical Journal Special Topics: Dynamics of
  Animal Systems}, vol. 224, no. 17--18, pp. 3141--3174, 2015.

\bibitem{Purcell:1977}
E.~M. Purcell, ``Life at low {R}eynolds numbers,'' \emph{American Journal of
  Physics}, vol.~45, no.~1, pp. 3--11, January 1977.

\bibitem{Hatton:2011IJRR}
R.~L. Hatton and H.~Choset, ``Geometric motion planning: The local connection,
  {S}tokes' theorem, and the importance of coordinate choice,''
  \emph{International Journal of Robotics Research}, vol.~30, no.~8, pp.
  988--1014, July 2011.

\bibitem{revzen2019stokes}
M.~D. Kvalheim, B.~Bittner, and S.~Revzen, ``\BIBforeignlanguage{English}{Gait
  modeling and optimization for the perturbed stokes regime},''
  \emph{\BIBforeignlanguage{English}{Nonlinear Dynamics}}, vol.~97, no.~4, pp.
  2249--2270, 09 2019.

\bibitem{ramasamy:thesis}
S.~Ramasamy, ``Geometry of locomotion,'' Ph.D. dissertation, MIME, Oregon St.
  Univ., Corvallis OR, 2020.

\end{thebibliography}

\end{document}